\documentclass[letterpaper, 10 pt, conference]{ieeeconf}  

\IEEEoverridecommandlockouts                              

\usepackage{graphics} 
\usepackage{caption}
\usepackage[space]{cite}
\usepackage{epsfig} 
\usepackage{amsmath} 
\usepackage{amssymb}  
\usepackage{bm}
\usepackage{algorithm,algcompatible,algpseudocode} 
\usepackage{xcolor}
\usepackage{multirow}

\usepackage{float}
\newcommand\blfootnote[1]{%
  \begingroup
  \renewcommand\thefootnote{}\footnote{\scriptsize#1}%
  \addtocounter{footnote}{-1}%
  \endgroup
}

\title{\LARGE \bf
Whisker-Inspired Tactile Sensing for Contact Localization on Robot Manipulators
}

\author{Michael A. Lin$^{1}$, Emilio Reyes$^{1}$, Jeannette Bohg$^{2}$ and Mark R. Cutkosky$^{1}$
\thanks{$^{1}$M. Lin, E. Reyes and M. Cutkosky are with the Department of Mechanical Engineering, Stanford University, Stanford, CA 94305, USA
{\tt\small [mlinyang,ereyes35,cutkosky]@stanford.edu}}
\thanks{$^{2}$J. Bohg is with the Department of Computer Science, Stanford University, Stanford, CA 94305, USA
{\tt\small bohg@stanford.edu}}
\thanks{Toyota Research Institute (TRI) provided funds to assist the authors with their research but this article solely reflects the opinions and conclusions of its authors and not TRI or any other Toyota entity.}
}

\begin{document}
\maketitle
\thispagestyle{empty}
\pagestyle{empty}

\begin{abstract}
Perceiving the environment through touch is important for robots to reach in cluttered environments, but devising a way to sense without disturbing objects is challenging. This work presents the design and modelling of whisker-inspired sensors that attach to the surface of a robot manipulator to sense its surrounding through light contacts. We obtain a sensor model using a calibration process that applies to straight and curved whiskers. We then propose a sensing algorithm using Bayesian filtering to localize contact points. The algorithm combines the accurate proprioceptive sensing of the robot and sensor readings from the deflections of the whiskers. Our results show that our algorithm is able to track contact points with sub-millimeter accuracy, outperforming a baseline method. Finally, we demonstrate our sensor and perception method in a real-world system where a robot moves in between free-standing objects and uses the whisker sensors to track contacts tracing object contours.
\end{abstract}

\blfootnote{\copyright 2022 IEEE. Personal use of this material is permitted.  Permission from IEEE must be obtained for all other uses, in any current or future media, including reprinting/republishing this material for advertising or promotional purposes, creating new collective works, for resale or redistribution to servers or lists, or reuse of any copyrighted component of this work in other works.}
\section{INTRODUCTION}
Manipulation in unstructured and cluttered environments is characterized by limited visibility and constrained motion. To improve robot perception, it is advantageous to sense the environment through multiple modalities including touch~\cite{kemp2007challenges}. Reaching into a cupboard or a refrigerator full of objects to retrieve a particular object are examples of tasks where contacts with objects may happen frequently and unexpectedly -- not only at the end-effector but along the entire arm. Sensing the locations and forces of contacts has been recognized as helpful for perception \cite{jain2013reaching,wang2020contact,bohg2010strategies,petrovskaya2011global} enabling access to locations that may be occluded from visual sensors. However, sensing through contact when interacting with free-standing objects that are small and light (e.g. a nearly empty bottle of pills or spices) is challenging because the action of making contact will likely change the object's state. Prior work has shown that minimizing the robot's inertial properties at the end-effector allows non-disturbing contacts and facilitates environment sensing \cite{bhatia2019direct,wang2020contact,lin2021exploratory}. However, a different approach is needed to detect and minimize disturbances arising from contacts along the robot arm.

\begin{figure}[thpb!]
\vspace{4 pt}
  \centering
  \includegraphics[width=\linewidth]{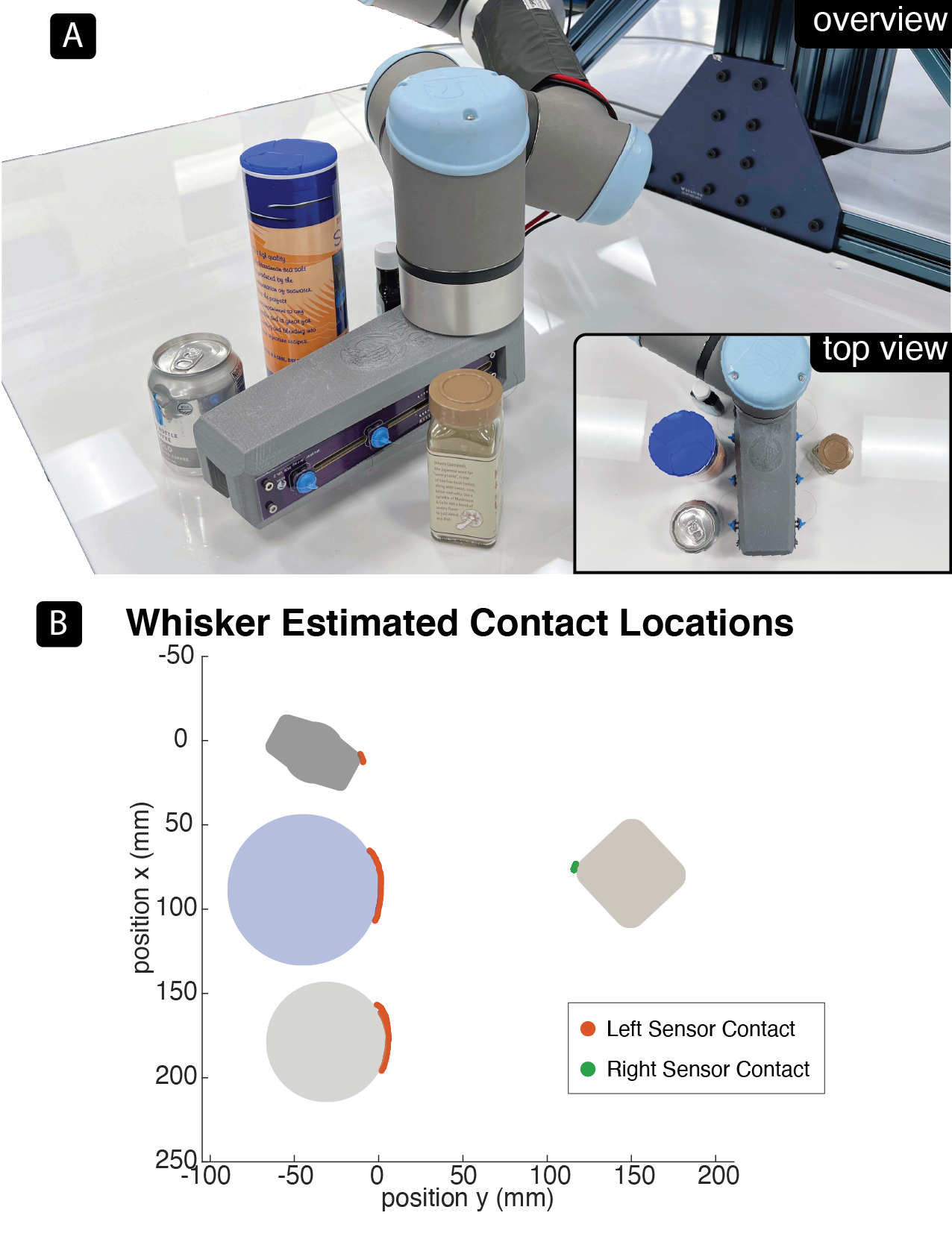}
  \caption{A) Demonstration of contact localization with two curved whisker sensor arrays on a robot arm reaching among objects of unknown shape and location. B) Plots of left and right sensor estimated contact locations. A video of this experiment is attached.}
  \label{fig:system}
  \vspace{-2em}
\end{figure}

In nature, we encounter various solutions for non-intrusive contact sensing. A common method is to use whiskers or vibrissae to navigate environments or localize prey in conditions where vision does not suffice~\cite{carvell1990biometric,catania2019mole}. In some cases (e.g. rats) animals employ sophisticated active whisking but in others, such as the vibrissae on the lower limbs of cats, whiskers are mostly passive and provide information while negotiating complex environments. The latter serves as the primary inspiration for the approach presented here.


Although whiskers are not widely used in robotics, there are a number of notable examples including \cite{prescott2009whisking,kaneko1998active,kim2007biomimetic,solomon2010extracting}. Most of these involve active whisking with a rotational joint and actuator at the whisker base. Passive whiskers (or antennae) on mobile bases are less common although there are a few examples (e.g., \cite{cowan2005biologically,kent2021whisksight}). In most of these cases, the whiskers are relatively stiff and are often straight, which means that if they approach a very light object head-on, they are likely to disturb it.

In our approach, we use very soft whiskers that bend easily and may be pre-curved to minimize disturbances and avoid the highly nonlinear phenomenon of buckling for slender elastic columns. The whiskers are mounted along a robot arm and we use the robot's accurate proprioception in combination with sensor measurements to localize contacts along the whisker and gather information about the environment.

A number of notable challenges arise in passive whisker deployment on a robot arm. The first is that whisker motion and deflection are subject to arm motion, which is typically intended to control a distal end-effector and not to provide exploratory sensing. Interaction with objects will often be limited to one sustained contact as opposed to multiple probing actions. The state estimation method should be able to process this arbitrary motion and whisker deflection to infer contact locations quickly. A second challenge is that as a whisker sensor is moved in arbitrary directions a straight whisker may catch its tip on object surfaces and buckle, making the sensor signal difficult to interpret. A sensor design with a curved whisker, as shown in Fig.~\ref{fig:signal}A, can avoid this buckling effect but contact localization using these whisker geometry may be more challenging and have been less explored.

\textbf{Contributions}: We present a new sensor design and fabrication method for creating arrays of slender, curved super-elastic nitinol vibrissae or whiskers mounted along the arm of a robot. We present a calibration method applicable to curved whiskers and a Bayesian filtering algorithm that can quickly track contact locations to within sub-millimeter accuracy. Then we implement three different Bayesian filters (Extended Kalman Filter, Unscented Kalman Filter and Particle Filter), showing that these methods can perform better than a baseline method \cite{solomon2010extracting}, with the UKF being the best performing filter in tests. Finally, we mount the sensors on a robot arm and demonstrate the ability to combine robot proprioception and sensor measurements to accurately track contact locations over time, allowing the arm to maneuver safely without disturbing even small and lightweight objects.

\section{RELATED WORK ON CONTACT INTERPRETATION}
\label{sec:related}
Although the sensing method presented here is new, it builds upon prior work on perception of unstructured environments through contacts \cite{petrovskaya2011global,bohg2010strategies,wang2020contact,koval2015pose,manuelli2016localizing,suresh2020tactile,lin2021exploratory}. A common challenge when sensing free-standing objects is that the act of contact sensing often will change the state of the object. For objects of known shapes, Koval \emph{et al.} developed a Particle Filtering approach to estimate object location through a sequence of pushes to collapse the belief distribution \cite{koval2015pose}. Suresh \emph{et al.} expanded on this work by posing the problem as a Simultaneous Localization And Mapping (SLAM) problem and were able to estimate both shape and location of objects \cite{suresh2020tactile}. While these methods work when interacting with isolated objects, the pushing approach is more challenging when an object is amidst clutter which constrains both the object and robot arm.

Sensing through non-intrusive contacts, on the other hand, has the advantage of objects remaining static, making state estimation easier. The most common of such perception methods is vision, however, RGB-D cameras are not well suited for close range sensing as is typically necessary when reaching into confined spaces with objects. Some work has investigated using close range proximity sensing \cite{hsiao2009proximity,tsuji2019proximity,schlegl2013virtual}, but these methods do not perform well when sensing specular or transparent surfaces for optical transducers, or may be susceptible to variations in materials properties for magnetic and capacitive transducers. Sensing through mechanical contact is not affected by these problems.

As noted earlier, a modest number of investigations have addressed whisker- or antenna-based sensing in robotics. Early work by Kaneko \emph{et al.} showed a method of active probing where a flexible antenna is rotated by actuators at one end to make contact with objects while estimating contact location with measured rotational compliance \cite{kaneko1998active}. Subsequent efforts addressed improved contact localization for objects of varying shapes \cite{solomon2010extracting,kim2007biomimetic,merker2021vibrissa}. Using 3-DOF force/torque sensing at the base (two bending torques and one axial force) it is possible to deduce contact locations from single measurements \cite{huet2017tactile,emnett2018novel,nguyen2020contact}. However, these methods either require using additional actuators for whisking or, in the latter case, require complex models to fit a unique mapping and accuracy is limited in practice.



\begin{figure}[!htpb]
    \vspace{-1em}
  \centering
  \includegraphics[width=\linewidth]{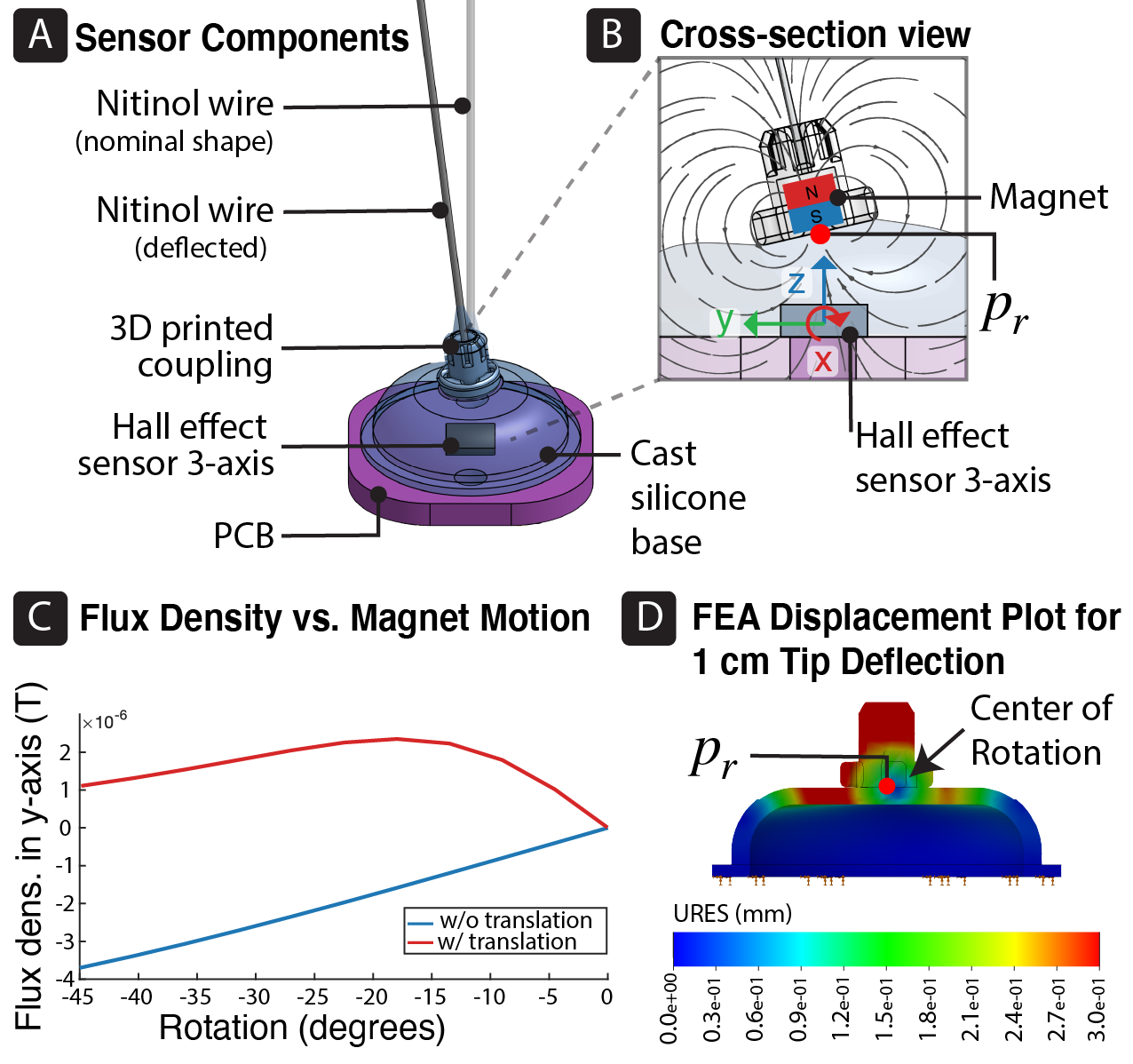}
  \caption{
  A) Sensor design components including flexible nitinol wire and compliant base. B) Cross-section view of the base. Whisker deflection results in magnet rotation around point $p_r$ and a change in magnetic flux measured by the Hall effect sensor. C) Minimizing rotation and translation coupling of the magnet yields a nearly linear sensor reading (blue line). D) FEA of the sensor as tip is displaced laterally by 
  1\,cm. Whisker deflections primarily result in magnet rotation.}
    \vspace{-1em}
  \label{fig:sensor}
\end{figure}

\section{SENSOR DESIGN AND FABRICATION}


\subsection{Overview}
Fig.~\ref{fig:sensor}A shows an overview of our sensor design and components. The sensor consists of a thin super-elastic nitinol wire bonded to a compliant silicone base that senses bending moment on the wire. Two variations of our sensor design are used in this work, one uses a straight nitinol wire (0.2\,mm diameter and 55\,mm long) and a second one uses a custom-shaped curved wire (0.2\,mm diameter, 20 mm in arc radius and 60\,mm in arc length). These wires are attached at one end to a neodymium permanent magnet through a 3D printed coupling. The magnet is axially magnetized with a field direction coinciding with the axis. The magnet and wire are supported by a hollow dome of silicone rubber that promotes rotation over translation as the wire is loaded in contact. The dome is centered over a triaxial Hall effect sensor (Melexis MLX90393) mounted to a PCB.

Contacts along the whisker produce rotations at the base, which change the magnetic flux measured by the Hall effect sensor (Fig.~\ref{fig:sensor}B). The transduction principle is similar to that in \cite{kim2019magnetically} but with substantially lower stiffness (0.17\,mNm/rad v.s. 3.2\,mNm/rad) and designed for incorporation into an array.

We performed a COMSOL Multiphysics analysis to investigate the effects of magnet motion on the Hall effect sensor. As seen in Fig.~\ref{fig:sensor}C, the change in magnetic flux in the Y sensing axis is nearly linear when the motion of the magnet is pure rotation. In contrast, if the rotation is coupled with translation in the direction of applied force then the change in flux is non-linear and flux density change is substantially smaller. We therefore designed the compliant silicone base to favor rotational displacements over translation when the radial loads are applied to the wire. An advantage of this design is that it allows one to sense deflections even when using a very thin and flexible wire, which promotes non-intrusive sensing. Figure \ref{fig:sensor}D shows that the center of rotation (center of blue region indicating low displacement) is very close to point $P_r$.

The chosen sensor design parameters are: dome thickness: 0.6\,mm; dome height: 3.25\,mm; whisker length: 55\,mm. While we cannot claim they are optimal, they appear to provide a favorable combination of monotonic and approximately linear sensor response and large signal/noise ratio without saturation for anticipated contacts.

\subsection{Sensor Fabrication}
\begin{figure}[thpb]
\vspace{4 pt}
  \centering
  \includegraphics[width=\linewidth]{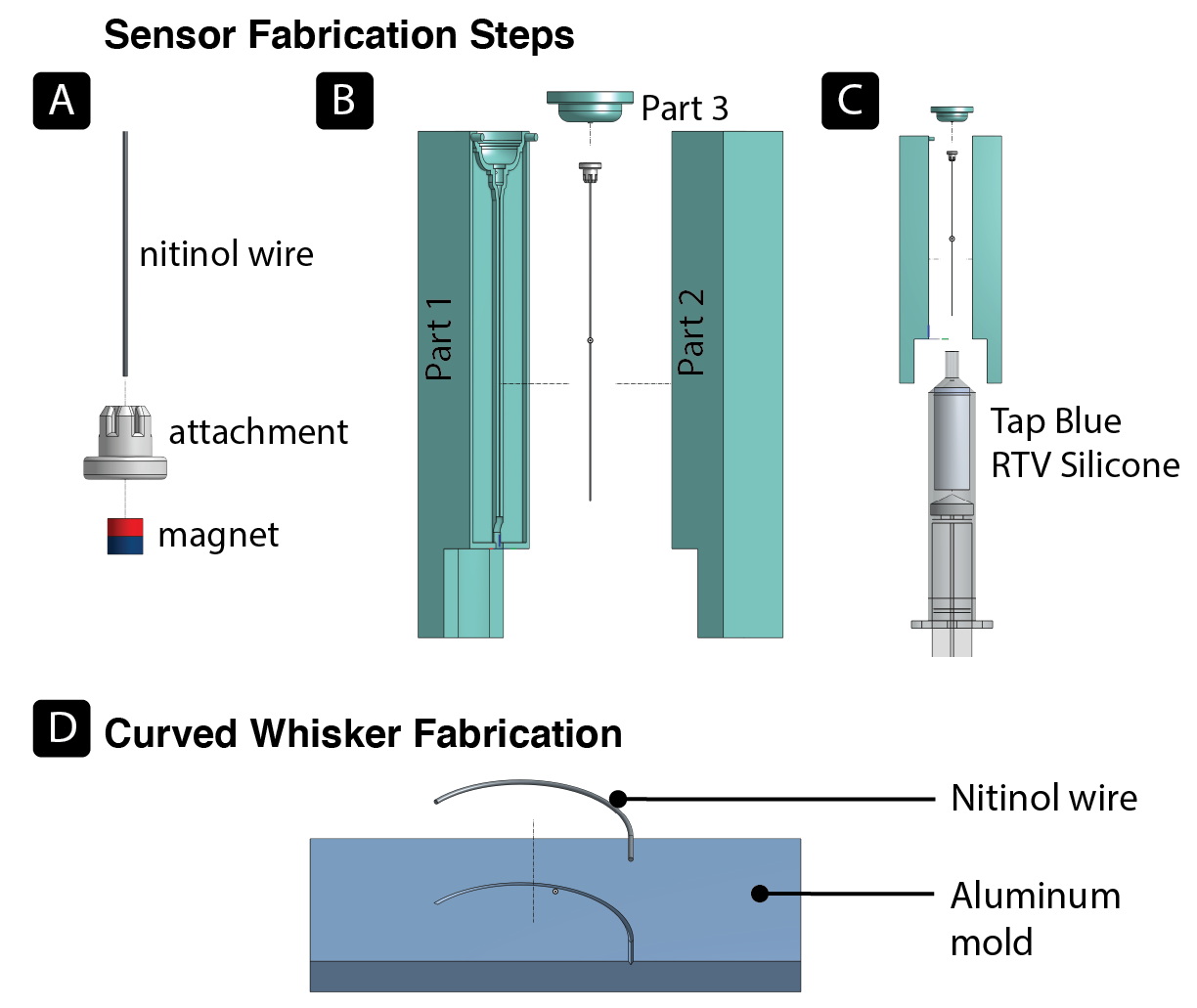}
  \caption{Whisker Sensor Fabrication. A) nitinol wire and magnet are glued to coupling piece. B) Whisker is clamped in 3-Part mold. C) Silicone is injected. D) Curved whiskers are pre-formed using heat.}
  \vspace{-1em}
  \label{fig:fab}
\end{figure}

\subsubsection{Sensor}
Whisker sensors are fabricated with a three part 3D printed mold. The molds and whisker/magnet coupling piece were printed with a high resolution SLA printer (Stratasys, Objet 30). The overall fabrication process is illustrated in Fig.~\ref{fig:fab}. The first step is to glue the magnet and a 0.2\,mm dia. x 55\,mm long nitinol wire (ASTM F2063, Malin Co. Cleveland, OH) to the attachment piece (Fig.~\ref{fig:fab}A). We used an axially magnetized N-52 grade neodymium magnet (size 1.6\,mm dia. x 1.6\,mm tall) with a surface field of 6619 Gauss (D11-N52, KJ Magnetics). The second step is to place this assembly into clamped mold Parts 1 and 2 (Fig.~\ref{fig:fab}B). The last step is to inject silicone (TAP Blue RTV) into the molds against gravity (Fig.~\ref{fig:fab}C). Part 3 is then placed with a weight on top to push excess material out. We chose the silicone for its durometer, ease of casting, and relatively low hysteresis \cite{lotters1997mechanical}. 

\subsubsection{Curved Whiskers}
Curved whiskers were fabricated by heat-forming nitinol wires in aluminum molds (Fig.~\ref{fig:fab}D). The goal is to produce a desired curvature while maintaining the super-elastic properties. The process
is sensitive to temperature and timing as well as wire size \cite{gilbert2015rapid}. We used a controlled oven at $600^\circ$C, placing the clamped molds with the nitinol wire and an embedded thermocouple. The temperature was monitored until it reached $500^\circ$C, at which point the setup was quenched in a water bath. The process keeps the metal for no more than 4.5 minutes at temperatures above $450^\circ$C to avoid aging and raising the austenite finish temperature.

\begin{figure}[thpb]
    \vspace{4pt}
  \centering
  \includegraphics[width=\linewidth]{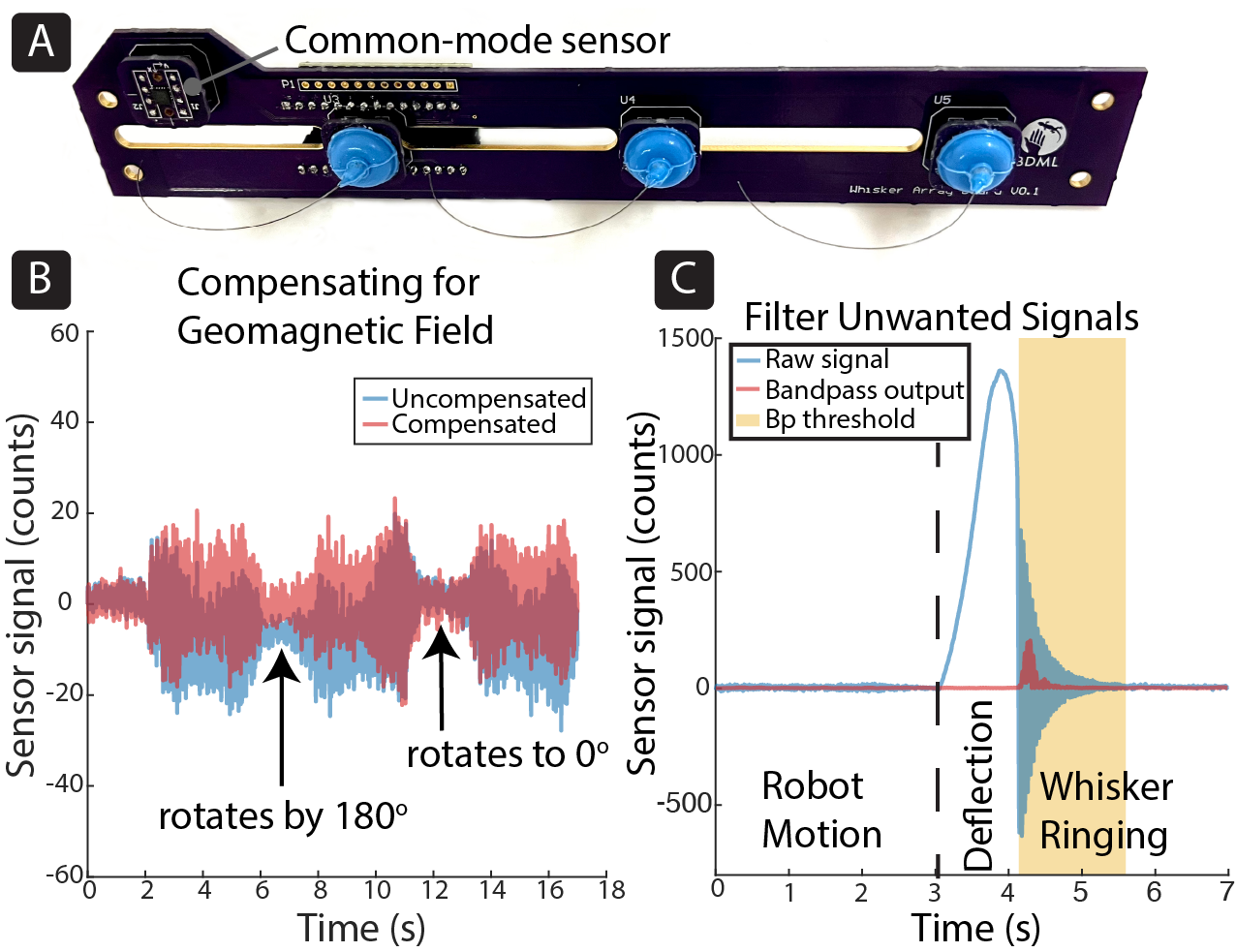}
  \caption{A) Sensor array with three curved whiskers and a common-mode sensor to compensate for geomagnetic field. B) Geomagnetic-compensated and uncompensated signals from one whisker as it is rotated $180^\circ$ in the direction of gravity. C) Time-series sensor signal when attached to a UR robot and moved in arbitrary motions in free-space. Ringing is detected by filtering.}
  \vspace{-1em}
  \label{fig:signal}
\end{figure}

\subsection{System Integration and Signal Conditioning}
The Hall effect sensor (MLX90393) is sampled at 250\,Hz through SPI communication with a Teensy 4.0 (ARM Cortex-M7 at 600 MHz). It is possible to sample an array of 20 sensors (3-axis each) at this frequency. Data arrays are sent to a computer through USB serial communication at the same rate. We also designed each array with a common-mode reference (Fig.~\ref{fig:signal}A) to compensate for the geomagnetic field.

In order to determine how to filter the sensors we performed an experiment with sensors mounted on a link attached to a Universal Robot UR10e arm. The link is commanded to move the sensor in free-space following an arbitrary trajectory -- in part, to test the effects of induced vibrations from robot actuators. At the end of the trajectory, the whisker is moved to make contact with an object causing a large deflection and later breaking contact, causing the whisker to return to its nominal shape, as it would happen in a use-case. Figure \ref{fig:signal} shows a typical time series plot of sensor signal, with small vibrations due to robot motion, followed by a large increase due to contact and a resonant vibration as the whisker breaks contact. Vibrations from robot motion can be mostly ignored as they account for less than 3\% of the maximum sensor signal. The ringing oscillation on breaking contact is due to the nitinol wire, and occurred at 26 Hz. We applied a 6th order Butterworth band-pass IIR filter with a passband frequency of 20 to 32 Hz, and used a threshold of the filtered output to detect these lost-contact events. 

\section{CONTACT LOCALIZATION}
\begin{figure}[thpb]
    \vspace{4pt}
  \centering
  \includegraphics[width=0.9\linewidth]{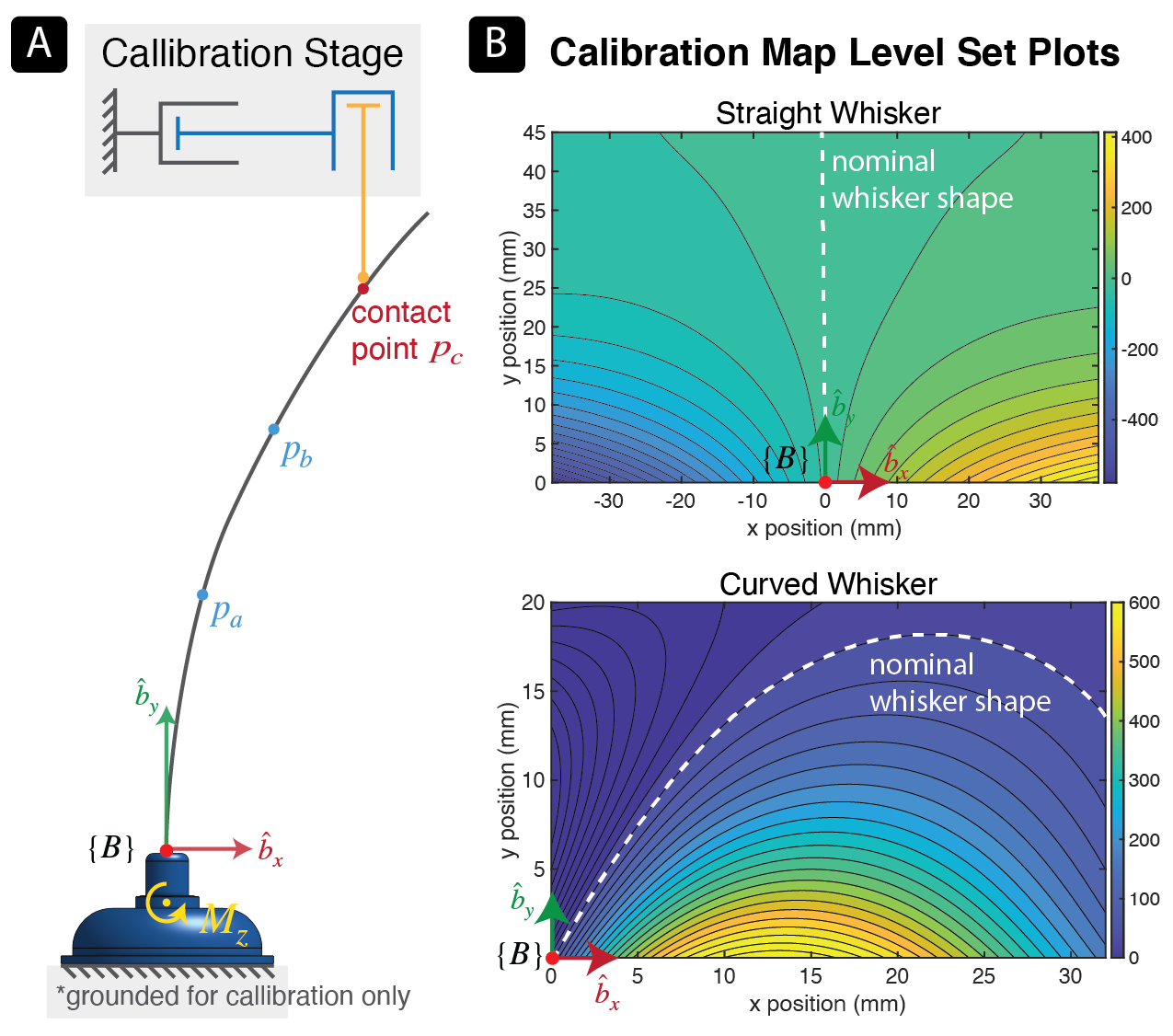}
  \caption{A) A whisker under deflection at contact point $p_c$ causes a bending moment ($M_z$) at the base. Contact at other points along the whisker, such as $p_a$ and $p_b$, result in the same bending moment. The sensor model $g$ that specifically maps $p_c$ to a measured $M_z$ is found through calibration with a 2-DOF prismatic joint stage. B) Results of calibration for both straight and curved whiskers are shown as contour plots of the 5th order polynomial models.}
  \vspace{-1em}
  \label{fig:formulation}
\end{figure}

To obtain information on the contact location along the whisker, we integrate instantaneous bending rotations at the whisker base due to whisker deflection, as well as motion of the whisker base over time. Base bending rotations are proportional to the whisker bending moment. However, there are many contact locations that can result in the same whisker deflection and, thus, moment at the base. With just instantaneous moment measurements, contacts at points $p_c$, $p_a$ and $p_b$ in Fig.~\ref{fig:formulation}A, would be indistinguishable.
Thus, in addition to instantaneous base moment measurements, we track the motion of the whisker base over time. In this section we detail our approach using these measurements and Bayesian filtering algorithms to infer contact point location and motion.

\subsection{Contact Localization Model and Assumptions} \label{sec:contactModel}

We begin by summarizing the assumptions made to limit the scope of the problem.
\begin{enumerate}
\item Objects that come into contact are immobile in the world reference frame. In other words, if the robot does not move then contact locations do not change.
\item There is at most one contact point on a whisker with the environment at any time. This will be true for convex objects.
\item Frictional forces along the whisker are negligible in terms of their effect on the sensor. This again will generally be true given the nitinol whisker material.
\item Motion of the robot as well as deflections of whiskers are in a 2D plane. This is generally true for objects that are flat in the direction of gravity, which we limit to in this work.
\end{enumerate}

In the following derivation we follow the conventions in Murray \emph{et al.}~\cite{murray2017mathematical} with lowercase boldface fonts for vectors, uppercase boldface fonts for matrices and $\{F\}$ denoting a reference frame F. Lowercase letters also refer to reference frames. For example, $\bm{v}^a_{ab}$ is the linear velocity of reference frame A relative to B (subscript ${ab}$) as viewed in reference frame A (super-script $a$).

Let $\{B\}$ be the reference frame of the sensor base and $\{S\}$ be the spatial (world-fixed) reference frame. We define our process and sensor model as follows
\begin{align}
    \bm{x}_{k+1} &= \bm{A}\bm{x}_k+\bm{B}\begin{bmatrix}\bm{v}^b_{sb}\\\bm{\omega}^b_{sb}\end{bmatrix}+\bm{w}_k\label{eq:procModel}\\
    y_k &= g(\bm{x}_k)+\bm{\nu}_k
\end{align}
Where $\bm{x}_k=\bm{p}_c$ is the position vector of the contact point relative to the origin of $\{B\}$ at time-step $k$. $y_k$ is the sensor measurement at time-step $k$. $\bm{v}^b_{sb}$ and $\bm{\omega}^b_{sb}$ are the linear and angular velocity, respectively, of sensor base $\{B\}$ relative to world-fixed frame $\{S\}$ as viewed in the body frame. Process and sensor noise are modelled as Gaussian white noise where $\bm{w}_k, \bm{\nu}_k\sim \mathcal{N}(0,\,\sigma^{2})$. $g:\mathbb{R}^2 \to \mathbb{R}$ is our sensor model that maps contact location to moment measurements which we will elaborate on in Section \ref{sec:calib}.


To define the process model, we first find the velocity of point $p_c$ relative to $\{S\}$ as viewed in the body frame $\{B\}$ denoted as $\bm{v}^b_{sp_c}$. Let's consider $p_c$ as a point with respect to $\{B\}$, and define a point $b_p$ that is instantaneously coincident to $p_c$, but fixed in $\{B\}$. We can find the linear velocity of $p_c$ with respect to the spatial frame as
\begin{align}
    \bm{v}^b_{sp_c} &= \bm{v}^b_{sb_p} + \bm{v}^b_{bp_c}\\
                    &= \bm{v}^b_{sb} + \bm{\omega}^b_{sb}\times \bm{b}_p + \bm{v}^b_{bp_c}
\end{align}
where $\bm{v}^b_{sb_p}$ is the linear velocity of point $b_p$ relative to $\{S\}$ as viewed in the body frame, $\bm{v}^b_{sb}$ is the linear velocity of $\{B\}$ relative to $\{S\}$ as viewed in the body frame, $\bm{\omega}^b_{sb}$ is the angular velocity of $\{B\}$ with respect to $\{S\}$. $\bm{p}_c$ is the position vector of point $p_c$ relative to $\{B\}$'s origin.

Given our assumption that the contact point remains static in the spatial frame (i.e. $\bm{v}^b_{sp_c}=\bm{0}$), the contact point velocity relative to $\{B\}$ is
\begin{align}
    \bm{v}^b_{bp_c} &= -\bm{v}^b_{sb} + -\bm{\omega}^b_{sb}\times \bm{b}_p\\
                    &= -\begin{bmatrix}\bm{I} & [\bm{p}_c]\end{bmatrix}\begin{bmatrix}\bm{v}^b_{sb}\\\bm{\omega}^b_{sb}\end{bmatrix}
\end{align}
where $[\bm{p}_c]$ is the skew-symmetric matrix of vector $\bm{p}_c$ and $\bm{I}$ is the identity matrix. 

As the sensor is attached on a robot link, $\bm{v}^b_{sb}$ and $\bm{\omega}^b_{sb}$ can be found through 
\begin{align}
    \begin{bmatrix}
\bm{v}^b_{sb}\\\bm{w}^b_{sb}
\end{bmatrix}=\bm{J}^b_{sb}(\bm{q})\bm{\dot{q}}
\end{align}
where $\bm{J}_{sb}^b$ is the body Jacobian of $\{B\}$ relative to $\{S\}$, and q is the vector of manipulator joint angles.

With the velocity of the contact point with respect to $\{B\}$ defined, we can express the process model as
\begin{align}
    \bm{x}_{k+1} &= \bm{x}_k + \delta_t\bm{v}^b_{sp_c} = \bm{x}_k -\delta_t\begin{bmatrix}\bm{I} & [\bm{p}_c]\end{bmatrix}\bm{\xi}^b_{sb}+\bm{w}_k
\end{align}
which gives as $\bm{A}=I$ and $\bm{B}=-\delta_t\begin{bmatrix}\bm{I} & [\bm{p}_c]\end{bmatrix}$ in Eq.~\ref{eq:procModel}.






In the case when contacts are not static, such as when contact travel along the contour of large radius of curvature objects, the velocity of the contact point relative to the world $\bm{v}^b_{sp_c}$ is non-zero. However, since we assume no prior knowledge of the object shape and location, the contact velocity is unpredictable. As inferences are executed very fast (250\,Hz) contact points cannot move far between iterations, so we choose to make the static assumption and correct for errors using the sensor model.

\subsection{Sensor Model and Calibration} \label{sec:calib}
In order to find the sensor model, $g:\mathbb{R}^2 \to \mathbb{R}$, which maps contact point locations, $\bm{x}_k$, to predicted moments measurements, $y_k$), we develop a calibration platform and procedure. To gather data for this mapping we used a calibration stage composed of two orthogonal prismatic joints and a thin rod (steel dowel pin of 0.4\,mm diameter) attached vertically at the tip of the end-effector (illustrated schematically in Fig.~\ref{fig:formulation}A). The position of the dowel pin is measured with linear optical encoders (US Digital EM-2) at a resolution of 0.0064\,mm/count. During data collection, position and magnetic sensor data are recorded as the end-effector of the stage is driven to deflect the whisker with the dowel pin. This procedure is then repeated for different positions of the dowel pin by tracing an arbitrary trajectory that spans the sensing region. For straight whisker sensors, data are collected for contacts on the left and right side of the whisker. For curved whiskers, data are collected for contacts that deflect the whisker towards the center of the arc.

We used a 5th-order bivariate polynomial model to fit the calibration data for both straight and curved sensors. Fig.~\ref{fig:formulation}B shows level set plots of the models for both straight and curved whisker sensors. Polynomial regression for the straight whisker data resulted in a R-squared value of 0.9987 and Root-Mean Squared Error (RMSE) of 0.503 (max. signal of 32). Regression for the curved whisker data resulted in a R-squared value of 0.9895 and RMSE of 1.348 (max. signal of 82).

While our 2D sensor calibration model works for the scenarios we consider here, in some real world cases deflections of the whisker out of the plane may occur due to contact with surfaces that are not vertical. 
As discussed in Sec.~\ref{sec:future} we can extend the calibration to include such effects.

\subsection{Contact Location Inference with Bayesian Filtering} \label{sec:bayesian}
To track contact locations from a sequence of base moment measurements, we use Bayesian filtering in a recursive algorithm that infers the state distribution from a history of sensor data and control inputs. The posterior distribution is defined as
\begin{align}
    b(\bm{x}_t) &=p(\bm{x}_t \mid \bm{z}_{1:t}, \bm{u}_{1:t}) \\
    &= \eta \; p(\bm{z}_t\mid \bm{x}_t,\bm{u}_t)\int p(\bm{x}_t\mid \bm{x}_{t-1},\bm{u}_t)b(\bm{x}_{t-1})d\bm{x}_{t-1}
\end{align}
where $\eta$ is a normalization factor, $b(\bm{x}_{t-1})$ is the prior distribution, and $p(\bm{z}_t\mid \bm{x}_t,\bm{u}_t)$ and $p(\bm{x}_t\mid \bm{x}_{t-1},\bm{u}_t)$ can be obtained from the sensor and process model, respectively.

We implemented three different non-linear Bayesian filters to compare their performance: Extended Kalman Filter (EKF), Unscented Kalman Filter (UKF) and Particle Filter (PF). A process noise covariance of $1e^{-5}I_2$ was empirically found to work well for EKF and UKF. This low noise is reasonable given the accuracy of the optical encoders and calibration system. The sensor noise variance was chosen to be 0.25 which was found through the RMSE of the calibration results reported previously. The process and sensor noise covariance for PF $1e^{-3}I_2$ and 1 respectively, which were also found empirically to track with low error. We used N=1000 particles for PF.

As mentioned previously, when the actual contact location travels along the surface of an object, our process model described in Sec. \ref{sec:contactModel} will be inaccurate which may lead to divergence of the estimate. We cannot model this behavior, but we can use known techniques for compensating for model errors. These include adding fictitious process noise or using a Fading Memory (FM) filter \cite[p.\,139]{simon2006optimal}. Both are used to increase the predicted covariance, but do so differently. FM scales the prior covariance while fictitious process noise adds a constant positive variance to the diagonals. FM was empirically found to produce better results in our application. We implemented this method on the Bayesian filters by scaling the prior covariance by a factor $\alpha=1.004$ at every time-step. 


In addition to implementing these filters, we compared tracking performance to a baseline by Solomon \emph{et al.}~\cite{solomon2010extracting}. Similar to our method, this baseline estimates contact points at every time-step but with estimates that are deterministic rather than probabilistic (i.e. single values rather than a distribution). Another distinction between our method and this baseline is that our sensor model is calibrated, while the baseline uses a small deflection Euler-Bernoulli beam model. In our implementation of this baseline, we used our calibrated sensor model to obtain predicted moments ($M_{i+1}$) and the arc-length to torque ratio ($\frac{ds}{M_i}$) which are used for the estimation correction step. For more detail on the baseline please refer to \cite{solomon2010extracting}.


\subsection{Contact Localization Experiments}
To compare the tracking performance of all algorithms, we collected data from two experiments, one using straight whiskers and one using curved whiskers. Each experiment consisted of 10 trials of data collected at 250\,Hz using the 2-DOF calibration stage. In each trial, the whisker makes contact and stays in contact with the thin dowel pin as it traces an arbitrary trajectory. Duration of each trial was on average 15 seconds. 

Initial conditions used for localization were the same across all conditions. The initial prior mean $\mu_0$ was set to be a constant 5\,mm offset in the $\hat{b}_y$ direction (illustrated in Fig.~\ref{fig:formulation}) from the ground truth. This is done to replicate error in initial estimate of the contact point along the whisker, which is realistic in practice due to sensor noise, hysteresis, model error, etc. Offset in the $\hat{b}_x$ direction is less important as, for a straight whisker, it can be captured accurately at the instance contact is detected. For the Bayesian filters, the initial prior covariance matrix was set to be $Q = \sigma^2I_2$, where $\sigma^2=10$. This large initial variance is a conservative choice to indicate high uncertainty in the initial contact location estimate. Inference for all methods is executed at sensor sampling rate of 250\,Hz.

\subsection{Contact Localization Results}
\begin{figure}[thpb]
  \centering
  \includegraphics[width=\linewidth]{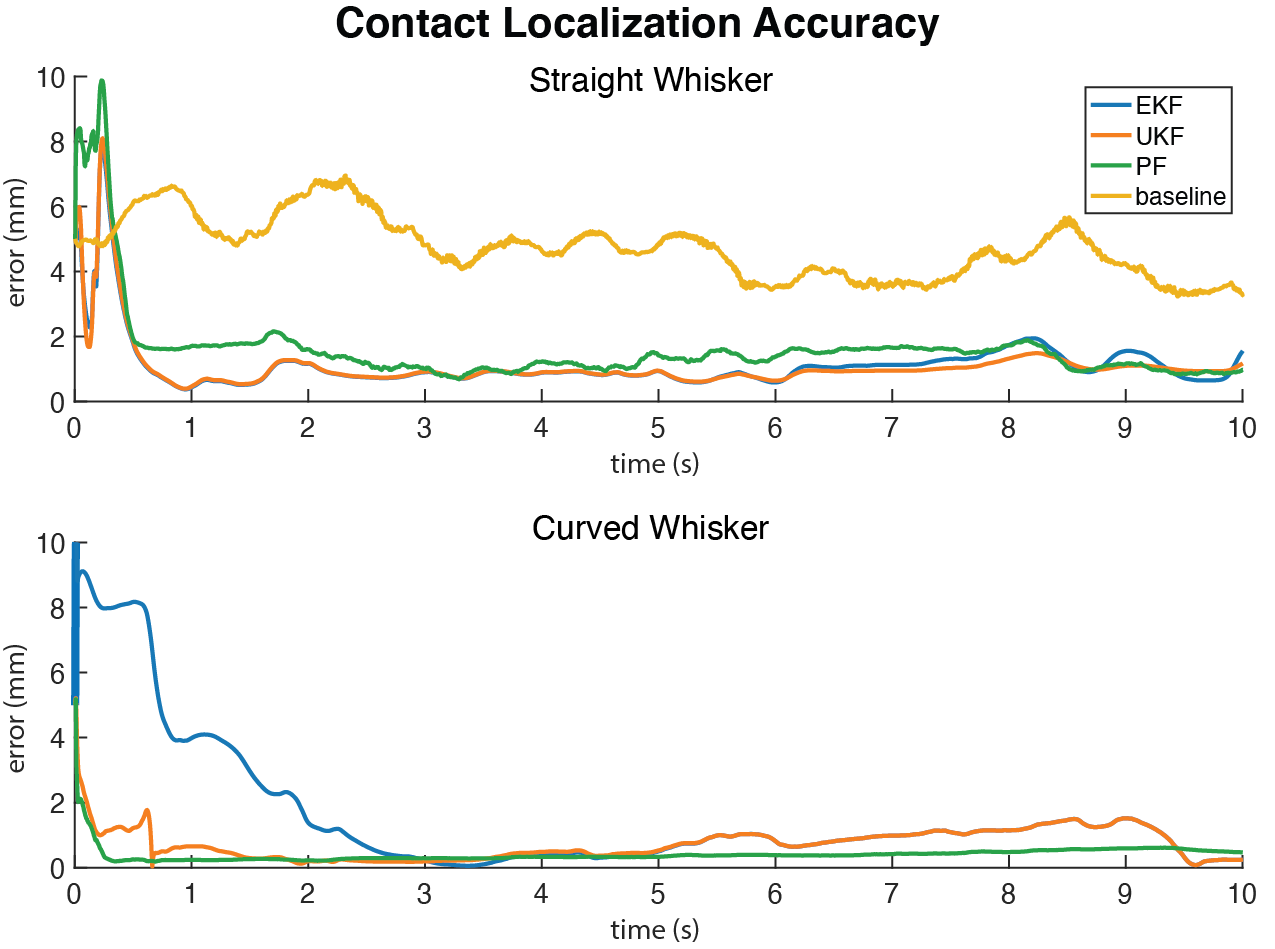}
  \caption{Plot of a single trial comparing contact localization accuracy among Bayesian filtering methods and baseline. Tracking errors for straight whisker sensor are shown in upper plot, errors for curved whiskers in the lower. Results from 10 trials are summarized in Table \ref{tab:res}}
  \label{fig:tracking}
\end{figure}

Results with the three Bayesian filtering and baseline methods are compared in Table \ref{tab:res} with tracking error over time for one trial shown in Fig.~\ref{fig:tracking}. While we show EKF, UKF and PF results for a straight and curved sensor, the baseline was executed only for the straight whisker case as this was an assumption in that model.

In the straight whisker experiments, all Bayesian filters tracked the ground truth contact location with a lower Euclidean distance compared to the baseline on average. UKF had the lowest average error of 0.838\,mm with a std. dev. of 0.603. In all trials, UKF converged to within 1\,mm of the true location within 0.7 seconds for both straight or curved whisker experiments. Fig.~\ref{fig:tracking} shows this convergence for one trial.

The results show that our Bayesian filtering methods are able to integrate sensor measurements and proprioceptive input to quickly infer contact location with relatively high accuracy, with EKF and UKF tracking with sub-millimeter accuracy on average. High accuracy tracking on both straight and curved whiskers shows that the Bayesian filtering method generalizes to both sensor designs even though they have different sensor models. These methods outperform the baseline which, to the authors' knowledge, is the closest work to that presented here. As mentioned in the cited paper, the baseline method requires an accurate estimate of the initial contact to perform well. However, in practice it is difficult to obtain accurate priors on contact location along the whisker. Some work has investigated using small angle deflection during initial contact to estimate initial location \cite{kaneko1998active} but this can still introduce errors due to object curvature, sensor noise or model errors. Obtaining good priors is even more challenging when using non-straight whiskers. Our curved whisker tracking results show that average errors are close to those in the straight whisker experiments. 

\begin{table}[h]
\vspace{5pt}
\caption{Tracking Performance Comparison}
\label{tab:res}
\begin{tabular}{|ll|l|l|l|l|}
\hline
\multicolumn{2}{|c|}{\multirow{2}{*}{Method}} & \multirow{2}{*}{\begin{tabular}[c]{@{}l@{}}Mean Dist.\\ (mm)\end{tabular}} & \multirow{2}{*}{\begin{tabular}[c]{@{}l@{}}Max. Dist.\\ (mm)\end{tabular}} & \multirow{2}{*}{\begin{tabular}[c]{@{}l@{}}Min. Dist.\\ (mm)\end{tabular}} & \multirow{2}{*}{\begin{tabular}[c]{@{}l@{}}Time\\ (ms)\end{tabular}} \\
\multicolumn{2}{|c|}{} &  &  &  &  \\ \hline
\multicolumn{1}{|l|}{\parbox[t]{2mm}{\multirow{4}{*}{\rotatebox[origin=c]{90}{Straight}}}} & baseline & 4.191 (SD 2.42) & 11.195 & 0.0214 & 0.533 \\ \cline{2-6} 
\multicolumn{1}{|l|}{} & EKF & 0.884 (SD 0.65) & 7.838 & 0.0222 & 0.346 \\ \cline{2-6} 
\multicolumn{1}{|l|}{} & UKF & 0.838 (SD 0.60) & 8.104 & 0.0101 & 0.565 \\ \cline{2-6} 
\multicolumn{1}{|l|}{} & PF & 1.302 (SD 0.99) & 9.877 & 0.0234 & 2.292 \\ \hline
\multicolumn{1}{|l|}{\multirow{3}{*}{\rotatebox[origin=c]{90}{Curved}}} & EKF & 0.896 (SD 1.17) & 10.048 & 0.0021 & 0.347 \\ \cline{2-6} 
\multicolumn{1}{|l|}{} & UKF & 0.617 (SD 0.55) & 5.543 & 0.0006 & 0.525 \\ \cline{2-6} 
\multicolumn{1}{|l|}{} & PF & 0.690 (SD 0.72) & 6.948 & 0.003 & 2.082 \\ \hline
\end{tabular}
\end{table}

\subsection{Contact Localization on Moving Contact Points}
To test the Bayesian filters on moving contacts, we collected data for contacts between an initially straight whisker and objects of different shapes and sizes. This experiment was performed with the same setup as the previous experiment, however, objects of different shapes were
moved into contact with the whisker, instead of the thin rod. In each trial, a whisker was deflected by bringing it into contact with an object. The whisker was moved laterally back and forth three times, such that the contact location traveled over the surface, tracing the contour of the object.

\begin{figure}[thpb]
  \centering
  \vspace{4pt}
  \includegraphics[width=\linewidth]{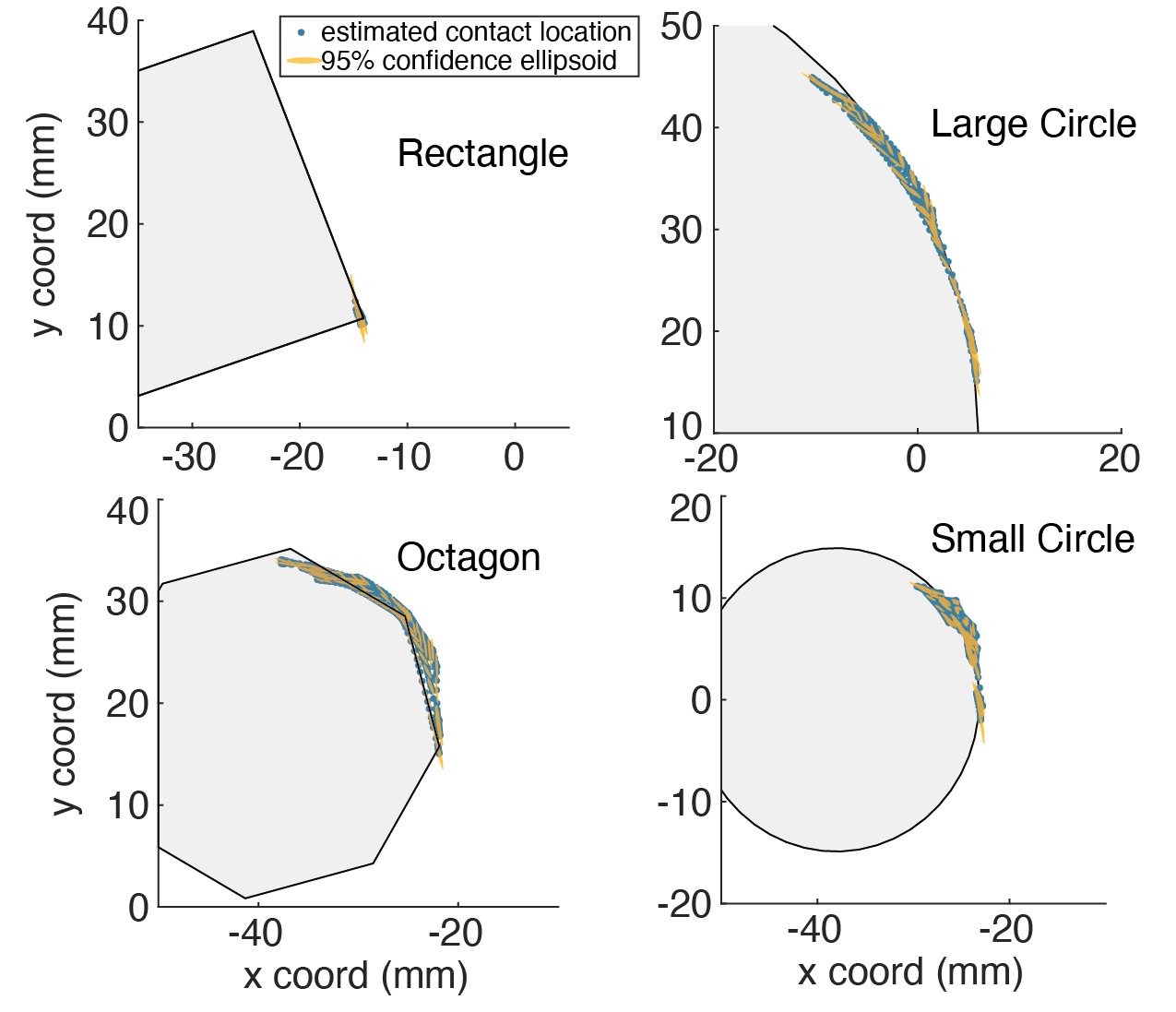}
  \caption{Tracking contact point location when making contact with objects of different shapes. Estimated contact points are downsampled in time by 20 for better visualization.}
  \label{fig:shapes}
\end{figure}

We selected the filter with the lowest mean error (UKF) to execute the contact localization for these experiments. Four objects were used: a rectangular prism (width: 30\,mm, length: 40\,mm), a small cylinder (diameter: 30\,mm), a large cylinder (diameter: 100\,mm) and an octagonal prism (side length: 
12.4\,mm). As ground truth we use the known location and shape of the object, allowing us to compare the contour predicted by our tracking algorithm. Tracking results can be visualized by how closely the contact location traces the object contour sweeping back and forth three times. Estimated contact locations are shown as a scatter plot in Fig.~\ref{fig:shapes} overlaid with the ground truth object contour. We computed the minimum distance of estimated contact points to the surface of each object contour. Average distance for each shape was: large cylinder = 0.460\,mm (std.~dev.~0.389), small cylinder = 0.378\,mm (std.~dev.~0.290), octagonal prism = 0.521\,mm 
(std.~dev.~0.436) and rectangular prism = 0.744\,mm 
(std.~dev.~0.670).

It can be seen in Fig.~\ref{fig:shapes} that, for the cylinders and octagonal prism, the filter produces three distinguishable sweeps along the object contour. For the case of the cuboid, the filter correctly estimates the contact location to remain on one corner. These results show that our tracking method is able to extract object contours in a real application and allow for object shape and location identification as a robot navigates an unstructured environment.

\section{SYSTEM DEMONSTRATION}
\label{sec:demonstration}
In this section we demonstrate the integration of arrays of whisker sensors with a Universal Robot UR16e robot arm (as shown in Fig.~\ref{fig:system}) in a task of reaching among objects of unknown location and shape while using our sensors and algorithms to continuously localize contacts. In this experiment, we use two curved whisker arrays that are attached to both sides of the end-effector, where each sensor is individually calibrated. The table setup included a heavy salt bottle, an aluminum can filled with liquid, a seasoning bottle and a light (60\,grams) vanilla extract bottle.

The arm is commanded to move the end-effector in a pre-planned trajectory such that the whiskers on both sides brush against the objects' surfaces. During this execution we record the pose of the end-effector and the sensor readings at a rate of 
250\,Hz using ROS. The data is then post-processed offline.

We used UKF to perform the contact localization. When contact is detected on a sensor, we initialize the Bayesian filter with a prior mean at coordinates near the sensor frame origin and the same prior covariance as specified in Section \ref{sec:bayesian}. After a sensor loses contact, these parameters are re-initialized. 

In Fig.~\ref{fig:system}B, we show the scatter plot of contact localization results from the two most distal sensors (referred to as left and right sensor). We are able to extract the local contours of the objects, which can be most clearly visualized with the cylindrical objects through only one sweep of the whisker. Moreover, throughout the reaching experiment, objects were undisturbed from their initial location as contact was made with the whisker. This shows that the robot is able to perceive the environment through non-intrusive contact.




\section{CONCLUSION \& FUTURE WORK}
\label{sec:future}
Robot tactile perception in unstructured environments can be enhanced by whisker sensors that sense through low-force and non-intrusive contacts. To make these sensors more effective, our design features low mechanical stiffness (0.17\,mNm/rad) such that the sensors can afford non-intrusive contact with light-weight free-standing objects as shown in Section \ref{sec:demonstration}. In addition, we developed a method to fabricate the sensor with a curved shape which facilitates passive ``whisking'' as a byproduct of robot motion. We formulate a method for inferring contact locations using Bayesian filtering which achieves an localization accuracy of 1\,mm within 0.7 seconds. This method not only improves upon prior methods, but also can be generalized to a curved whisker geometry. Our results show that the sensor and algorithm can enable robots to perceive local object shapes and contact locations as it navigates in proximity -- without disturbing objects.

In the future, we wish to improve on our perception algorithms by considering information from multiple sensors jointly. Our current approach only uses measurements from one sensor, but with multiple whiskers, many contacts can happen simultaneously on one object. This can be used to discriminate object shape and location faster than through tracking single contact points.

Additionally, we currently calibrate our sensors in a 2D plane, which was sufficient for the contact types we tested, but we plan to extend our calibration process to consider the 3D case and account for whisker deflections out of the plane. To extend this sensor calibration method to 3D contact points we can modify the calibration stage such that the whisker base is rotated around its axis in order to deflect the whisker in 3D. A higher dimensional model can be used to map these 3D contact points to 3-axis sensor readings.

We are also interested in using whiskers to control robot motion. The high compliance properties of whiskers make them great candidates as they can quickly detect fast approaching contacts and provide feedback to a controller to react accordingly while exerting little force. This has the potential to be used to improve controllers that are used for reaching and navigating cluttered spaces \cite{jain2013reaching} by making contact safer and increasing the sensing volume of a robot arm.





\section*{ACKNOWLEDGMENT}

This work is supported in part by TRI Global and the authors thank our collaborator Dr. Hongkai Dai as well as students Jun En Low and Rachel Thomasson for insightful discussions. We also thank Fredrik Solberg for his help with the nitinol heat shaping setup.


\bibliographystyle{unsrt}
\bibliography{citations}

\end{document}